\pdfoutput=1
\documentclass[11pt,table,dvipsnames]{article}

\usepackage[final]{acl}

\usepackage{times}
\usepackage{latexsym}

\usepackage[T1]{fontenc}
\usepackage[utf8]{inputenc}

\usepackage{microtype}

\usepackage{inconsolata}

\usepackage{graphicx}
\usepackage{arydshln}
\usepackage{tabularx}
\usepackage{booktabs}
\usepackage{multirow, makecell, caption}
\usepackage{multicol}
\usepackage{xcolor}
\usepackage{colortbl}
\usepackage{listings}

\definecolor{color1}{RGB}{255, 240, 158}
\definecolor{color2}{RGB}{210, 235, 180}

\title{MultiLingPoT: Enhancing Mathematical Reasoning with Multilingual Program Fine-tuning}

\author{Nianqi Li$^{1}$, Zujie Liang$^{2}$, Siyu Yuan$^{3}$, Jiaqing Liang$^{3}$, Feng Wei$^{2}$, Yanghua Xiao$^{1}$\thanks{Corresponding author.} \\
$^{1}$Shanghai Key Laboratory of Data Science, School of Computer Science, Fudan University\\
$^{2}$MYbank, Ant Group
$^{3}$School of Data Science, Fudan University \\
\texttt{nqli23@m.fudan.edu.cn, shawyh@fudan.edu.cn}
}

\begin{document}
\maketitle
\begin{abstract}
Program-of-Thought (PoT), which aims to use programming language instead of natural language as an intermediate step in reasoning, is an important way for LLMs to solve mathematical problems.
% By offloading computation to a code interpreter, LLMs can leverage their reasoning abilities while overcoming computational limitations.
Since different programming languages excel in different areas, it is natural to use the most suitable language for solving specific problems.
However, current PoT research only focuses on single language PoT, ignoring the differences between different programming languages.
Therefore, this paper proposes an multilingual program reasoning method, MultiLingPoT.
This method allows the model to answer questions using multiple programming languages by fine-tuning on multilingual data.
% Additionally, hybrid methods, categorized into prior and posterior, are employed, allowing the model to select the most suitable language for each problem.
Additionally, prior and posterior hybrid methods are used to help the model select the most suitable language for each problem.
Our experimental results show that the training of MultiLingPoT improves each program's mathematical reasoning by about 2.5\%.
Moreover, with proper mixing, the performance of MultiLingPoT can be further improved, achieving a 6\% increase compared to the single-language PoT with the data augmentation.\footnote{Resources of this paper can be found at \url{https://github.com/Nianqi-Li/MultiLingPoT}}
\end{abstract}

\section{Introduction}
\label{sec:intro}
\begin{figure}[t]
    \centering
    \includegraphics[width=1.0\linewidth]{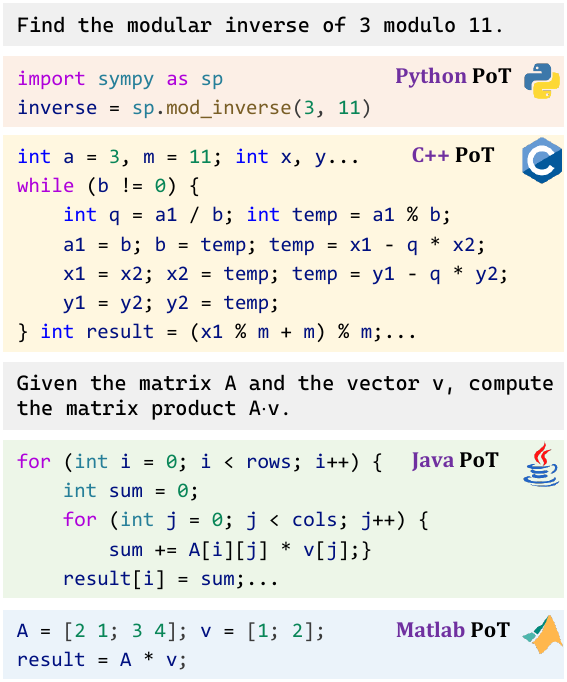}
    \caption{Examples of different programming languages having different advantages. For the given question, the suitable language is easy to answer while using other languages will be more difficult.}
    \label{fig:differentlanguagePoT}
\end{figure}

Program-of-Thought (PoT), which aims to use programming language instead of natural language as an intermediate step in reasoning in Large Language Models (LLMs), is an important way for LLMs to solve mathematical problems~\citep{chen2022program, gao2023pal}.
By generating programs and running them with a code interpreter, LLMs can not only exploit their reasoning capabilities but also avoid the computational errors that can be caused by using natural language.
% Significantly improve their mathematical reasoning ability.
Inspired by mathematical tasks, nowadays, PoT is also widely used in other domains, such as pseudo-code graph~\citep{skianis2024graph}, visual inference~\citep{suris2023vipergpt}, and document understanding~\cite{zhao2024docmath}.
Therefore, exploring the optimization of PoT is a problem of great value.

Previous work on PoT has concentrated on enhancing the mathematical capabilities based on a single language PoT, including techniques such as data augmentation~\citep{yue2023mammoth, jie2023leveraging}, PoT-CoT integration, and multi-round iteration~\citep{wang2023mathcoder, gou2023tora, qian2023creator}.
However, we believe that \textbf{different programming languages have different specializations}.
As shown in Figure~\ref{fig:differentlanguagePoT}, matlab is better at matrix operations than Java, while Python provides a rich library for number theory.
Therefore, it is a better approach to use the suitable language to solve the corresponding problem than to use one language to solve all problems.

Recently, \citet{luo2024python} also started to explore combining different programming languages.
However, their prompt-based approach is difficult to apply on small-scale models and does not explore how to choose the most appropriate programming language for a specific problem, which is the key to take advantage of multilingualism.
Besides, their comparison is biased due to the availability of more pre-trained data for multilingual models~\citep{li2024common}.
Therefore, how to enable small-parameter models to benefit from multilingualism and select the appropriate language for a problem remains an open question.

% Previous work on PoT has concentrated on data construction.
% Since \citet{chen2022program} proposed the program of thought prompting in 2022, research on PoT has received extensive attention.
% Constructing PoT data to fine-tune models becomes a common approach for enhancing mathematical abilities~\citep{yue2023mammoth, jie2023leveraging}.
% There are also studies aiming to combine program with natural language to enable multi-round thinking and computation~\citep{wang2023mathcoder, gou2023tora, qian2023creator}.
% However, there are obvious limitations in current PoT research:
% Most of the studies focus on PoT in a single programming language while ignoring the differences between different programming languages.
% As shown in Figure~\ref{fig:differentlanguagePoT}, different programming languages have different specializations, e.g. matlab specializes in matrix operations, while Python provides a rich library for number theory.
% Therefore, it is a better approach to use the suitable language to solve the corresponding problem than to use one language to solve all problems.
% \lnq{Recently, \citet{luo2024python} started to explore combining different programming languages.
% However, this research \lzj{only} focuses on prompt-based multi-program inference on the large-parameter model, which may be biased as multi-program involves more pre-training and ignores the potential of small-parameter models in this area.}

To fill the gap and further enhance the mathematical reasoning capability of LLMs, we introduce MultiLingPoT, a multilingual program reasoning method based on supervised fine-tuning.
This approach enables the model to solve mathematical problems using PoTs in multiple programming languages.
Specifically, given an input question, the model will generate an answer in the specified programming language.
To achieve this, we construct a large amount of multilingual programming data using ChatGPT~\citep{openai2022chatgpt} and remove samples with incorrect answers.
By fine-tuning on our high-quality multilingual PoT dataset, models can using the chosen programming language to answer mathematical questions.
% Given an input question, the model will generate an answer in the specified programming language.
Further, to adaptively select the most appropriate PoT for a given input, we incorporate prior and posterior hybrid strategies, including self-consistency, classifiers, and case-based choice, to mix results from different programs.
Finally, we conduct extensive experiments to validate the effectiveness of MultiLingPoT.
We find that the training of MultiLingPoT can improve the reasoning performance of the internal programming languages, achieving better results than single-language PoT by average of 2.5\%.
Even with data augmentation for single-language PoT, MultiLingPoT outperforms it by over 6\% with a suitable hybrid approach, and continues to show superior results compared to single-language PoT with self-consistency.

The main contributions of this paper are summarized as follows:
\begin{itemize}
    \item We construct simple and complex multilingual PoT datasets with 26,359 and 14,775 samples for Python, C++, Java, and Matlab.
    \item We propose an SFT-based multilingual program inference method, MultiLingPoT, and explore hybrid strategies that allow the model to answer the corresponding question using the most appropriate programming language.
    \item Through extensive experiments, we find that the training of MultiLingPoT enhances the reasoning ability of the internal languages. And with suitable mixing, MultiLingPoT can outperform single-language PoT with data augmentation by 6\% and still has the potential for further improvement.
\end{itemize}

\section{Related Work}
\label{sec:related}
\paragraph{Mathematical Reasoning}

Mathematical reasoning is an important metric to assess the ability of complex multi-step reasoning, which is a difficult task for neural networks~\citep{yang2023gpt}.
Recently, with the advent of LLMs, mathematical reasoning has improved greatly through methods such as finetuning~\citep{lewkowycz2022solving}, Chain-of-Thought~\citep{wei2022chain,wang2022self, zhou2022least, yu2023outcome, fu2023specializing, yu2023metamath, zhang2024self}.
% By fine-tuning on mathematical datasets, LLMs can significantly improve their complex reasoning capabilities~\citep{lewkowycz2022solving}.
% The introduction of CoT~\citep{wei2022chain}, which outputs intermediate steps, further boosts performance on mathematical datasets~\citep{wang2022self, zhou2022least, yu2023outcome, fu2023specializing, yu2023metamath, zhang2024self}.
However, \citet{gao2023pal} found that the performance of LLMs drops dramatically when dealing with complex computations~\citep{hendrycks2021measuring} or large numbers~\citep{geva2020injecting}.
Therefore, instead of requiring LLMs to use natural language for computation, \citet{gao2023pal} uses a code interpreter to help the model output results to avoid computational errors and become the primary method of mathematical reasoning today.
% By transferring the computation to the code interpreter, LLMs are able to avoid computational errors while exploiting reasoning capabilities, making program-based solutions the primary method of mathematical reasoning today.

\paragraph{Program-of-Thought}
Program-of-Thought (PoT), which aims to use programming language instead of natural language as an intermediate step in LLMs' reasoning, is an important way to solve mathematical problems.
In 2022, \citet{chen2022program} introduced the use of code as an intermediate step to assist LLMs, while \citet{gao2023pal} proposed the program-aided language model.
By building PoT data and fine-tuning, LLMs are able to enhance their mathematical capabilities~\citep{yue2023mammoth, jie2023leveraging, luo2023wizardmath, he2023solving}.
% Additionally, PoT combined with reinforcement learning~\citep{luo2023wizardmath} and symbolic solvers~\citep{he2023solving} have also been shown to improve the mathematical performance of the models.
Further, through iterations of thinking and program execution, models can combine CoT's reasoning with PoT's computation, resulting in models such as MathCoder~\citep{wang2023mathcoder} and ToRA~\citep{gou2023tora}.
However, these studies are limited to Python-based PoT, ignoring the differences between different programming languages.
Although, recent \citet{luo2024python} start to explore combining different programming languages.
Their prompt-based method is hard to apply on small-scale models, and lacks the exploration of choosing suitable language for specific question.
Basides, their comparsion is biased due to the larger pretraining volume for multilingual methods~\citep{li2024common}.
Therefore, this paper proposes an SFT-based multilingual program reasoning approach to select the appropriate language for the corresponding query, and explore the performance gains from multi-program fine-tuning with equal data volume.

\section{Training of MultiLingPoT}
\label{sec:training}
\begin{figure*}[t]
    \centering
    \includegraphics[width=0.98\linewidth]{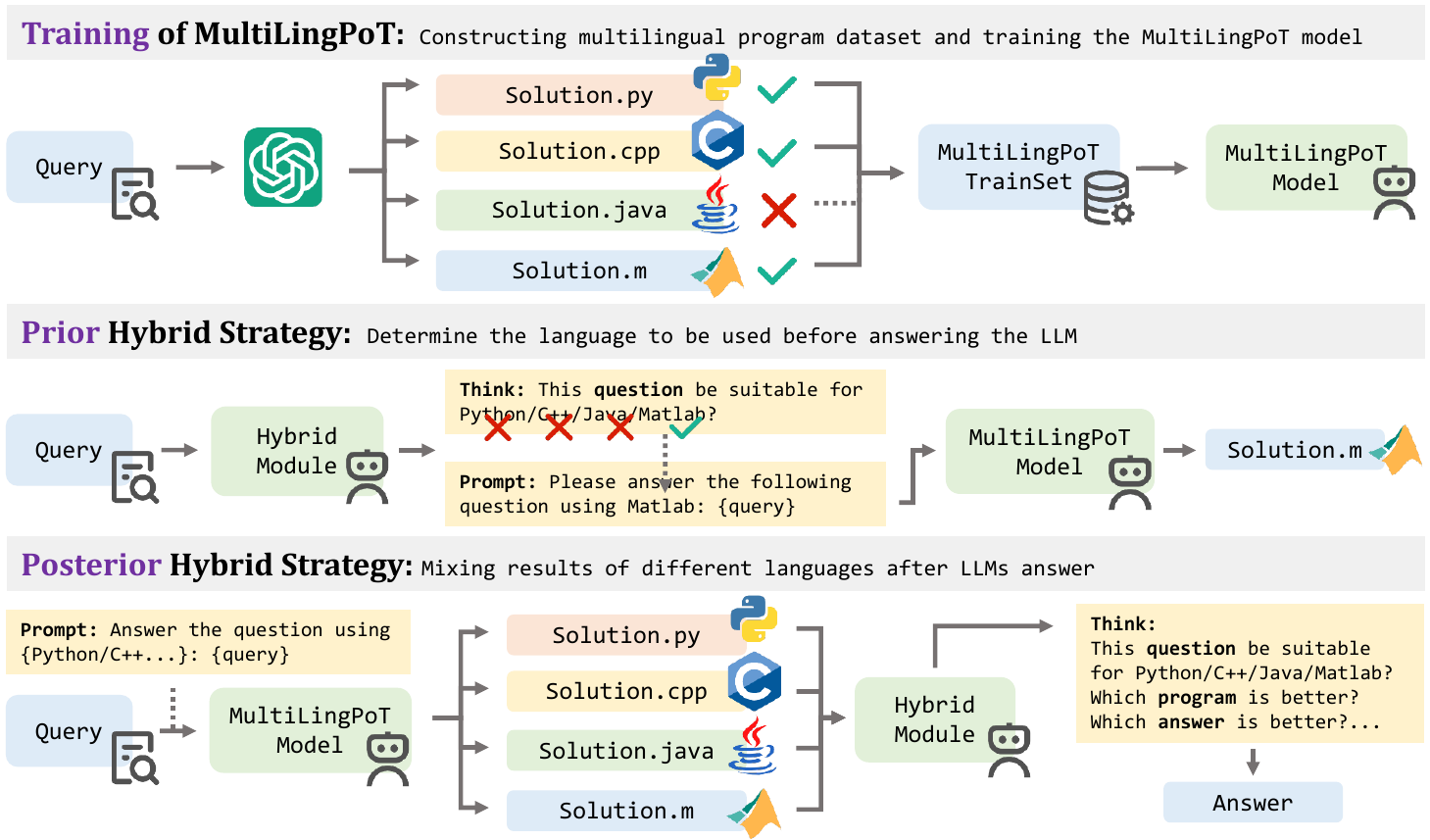}
    \caption{The illustration of the implementation of the MultiLingPoT methodology, including data construction, model training and the hybrid strategies. Considering the diverse implementations of hybrid strategies, the ``Think'' part only represents the underlying logic of the hybrid strategy, but not its specific implementation.}
    \label{fig:method}
\end{figure*}

In this section, we present the process of training MultiLingPoT model. 
The first part of Figure~\ref{fig:method} illustrates the implementation of this process.

\subsection{Language Selection}

To effectively implement and evaluate MultiLingPoT, it is crucial to select the appropriate programming languages.
Based on heuristic thinking, the chosen programming language should have the following characteristics:
1) The syntactic differences between languages should be significantly distinguishable so as to separate MultiLingPoT from single-language reasoning.
For example, C and C++ are not a good choice.
2) The selected languages should be popular, to avoid bias due to disparities in the model's proficiency across languages.
3) The language should support mathematical reasoning. 
For example, HTML, which is designed for the web, is not a suitable choice.

Based on the above criteria and following \citeposs{github2024octoverse} report, we choose Python, C++, Java, and Matlab for the experiments.

\subsection{MultiLingPoT Data Construction}
\label{sec:dataconstruct}

In order to teach the model to answer questions in multilingual programs, we construct training data across various languages.
Based on related work~\citep{jie2023leveraging, yue2023mammoth, luo2023wizardmath}, we select GSM8K~\citep{cobbe2021training} and MATH~\citep{hendrycks2021measuring} as the foundation for dataset construction, and used ChatGPT~\citep{openai2022chatgpt} to generate PoT data in multiple languages.
Specifically, for each problem in the trainsets, we instruct ChatGPT to generate solutions in four programming languages: Python, C++, Java, and Matlab.
To improve ChatGPT's ability to generate program in each language, we provide four manually crafted solution program as examples, which are provided in Appendix~\ref{appendix:data-construction}.
Due to the varying difficulty of GSM8K and MATH, different sets of examples are used for each, enabling better generation performance.
Finally, to ensure the quality of the training set, we execute all the generated programs using code interpreters.
Only the programs with correct results are kept.

Table~\ref{tab:data-construction} shows the results of our constructed dataset.
We collect 26,359 samples for GSM8K and 14,775 samples for MATH.
Since MATH is more challenging and has more erroneous outputs, the dataset for MATH is smaller than GSM8K.
However, as the data across different languages is balanced in both datasets, our data is suitable and fair for multilingual programming training.

\newcolumntype{a}{>{\centering\arraybackslash}p{0.75cm}}
\newcolumntype{b}{>{\columncolor{Orange!10}\centering\arraybackslash}p{0.75cm}}
\newcolumntype{d}{>{\columncolor{Yellow!10}\centering\arraybackslash}p{0.75cm}}
\newcolumntype{e}{>{\columncolor{Green!7}\centering\arraybackslash}p{0.75cm}}
\newcolumntype{f}{>{\columncolor{Blue!7}\centering\arraybackslash}p{0.75cm}}
\begin{table}[t]
  \centering
  \small
    \begin{tabular}{lbdddd}
    \toprule
    \rowcolor{white} \textbf{Dataset} & \textbf{Origin} & \textbf{Python} & \textbf{C++} & \textbf{Java} & \textbf{Matlab} \\
    \midrule
    GSM8K & 7473 & 6598 & 6535 & 6615 & 6612\\
    \cmidrule{2-6}
    MATH & 6282 & 3844 & 3737 & 3575 & 3619 \\
    \bottomrule
    \end{tabular}%
  \caption{Constructed PoT data for Python, C++, Java and Matlab based on GSM8K and MATH.}
  \label{tab:data-construction}%
\end{table}%

\subsection{MultiLingPoT Fine-Tuning}

Using the multilingual programming dataset, we train the MultiLingPoT model that can solve mathematical problems using different programming languages.
For an input query and a specified programming language, the MultiLingPoT model outputs a function in the corresponding language called ``solution'', which returns the result of the query.
For more details, the instruction and input templates are provided in Appendix~\ref{appendix:model-inference}.
Parameter settings and model selection for training are provided in Section~\ref{sec:trainsetup}.

\section{Hybrid Strategies in MultiLingPoT}
\label{sec:hybrid}
Since MultiLingPoT can answer questions using multiple programming languages, we explore hybrid strategies for  MultiLingPoT to select the most appropriate answer for input question.
Based on the timing of mixing, we categorize the hybrid strategies into prior and posterior.
The prior requires programming languages to be selected before LLMs answer, while the posterior allows LLMs to answer using all four languages and then mix them.
Figure~\ref{fig:method} shows the illustration of our hybrid strategies.

\subsection{Prior Hybrid Strategy}

The prior hybrid is a hybrid strategy only based on the query.
By selecting a language based on the query, this strategy determines the language to be used before the LLM generates the response, requiring only one generation and thus reducing computation.
Four specific implementations are explored under this strategy:

\paragraph{Case-Based Choice}
Assuming that similar queries have similar solutions~\citep{dong2022survey}, the method selects languages based on their performance on training examples similar to the input query.
For each input query, the method uses \texttt{text-embedding-3-small}~\citep{neelakantan2022text} to compute similarity and rank the relevant examples in the training set.
Starting with the most similar examples, the method counts the number of correct answers for each programming language.
The language that reaches 10 correct counts first is selected to answer the current query.

\paragraph{Small Model Scorer}
This method uses small parameter models as scorers for programming language selection.
Given that our task involves both natural and programming languages, we choose \texttt{Bert-base-uncased}~\citep{devlin2018bert} and \texttt{CodeBert-base}~\citep{feng2020codebert} as the base models.
Specifically, this method uses the four programming languages' performance on the training set to train four scorers.
Each scorer is responsible for evaluating a specific language.
For an input query, the scorers provide a rating of [0,1].
The language with the highest rating is chosen to answer the current query.

\paragraph{LLM Scorer}
Given that LLMs have better reasoning and understanding abilities, this method uses \texttt{Llama3-8B-Instruct}~\citep{llama3modelcard} as a scorer for programming language selection.
Similar to the small model scorer, this method trains a Llama3 scorer using the performance of four languages on the training set.
Given the input query and chosen language, the scorer returns a ``Yes'' or ``No'' response.
And we use the difference between the logprob of ``Yes'' and ``No'' as the score and select the highest scoring language to answer the current query.

\paragraph{Direct Perference Optimization}
This method performs DPO training~\citep{rafailov2024direct} on the MultiLingPoT model, giving the model the ability to select preferences itself.
Specifically, the method uses the performance of the four languages on the training set as the preference dataset.
For the input query, the PoT of the language that answered correctly is chosen, while the PoT of the incorrect language is rejected.
By DPO training on the preference dataset, the model can directly output the query-preferred language results without additional selection.

\subsection{Posterior Hybrid Strategy}

The posterior hybrid strategy is is a hybrid strategy based on the query and generated code in four languages.
Compared to the prior hybrid, the posterior hybrid has more information and better accuracy.
However, since each query requires four rounds of inference to generate answers in different programming languages, the posterior inference requires more computation time.
Four specific implementations are explored under this strategy:

\paragraph{Self Consistency}
Referring to the study by \citet{wang2022self}, this method selects the final answer by voting on the PoT results in four languages.
The answer with the most votes is chosen as the final answer.
In case of a tie vote, one of the tied results is randomly selected as the final answer.

\paragraph{Small Model Scorer}
This method uses small parameter models as the scorers for programming language selection.
All steps are the same as in the ``Small Model Scorer'' with the prior hybrid, except that both the query and the generated code are entered as criteria during training and inference.
This allows the model to make selections not only based on the preferences from the query but also by considering additional factors, such as the completeness of the code, the library functions used, etc.

\paragraph{LLM Scorer}
This method uses LLM as a scorer for programming language selection.
All steps are the same as in the ``LLM Scorer'' with the prior hybrid, except that both the query and the generated code are used as evaluation criteria.

\paragraph{Voting and Scoring}
Influenced by the fact that votes are often tied in complex dataset, we incorporate the self-consistency and LLM scorer methods.
For the input query and the PoT results in the four languages, the method first votes on the PoT results.
If there is a single highest-scoring answer, it is selected as the final output.
Otherwise, the final answer is selected using the LLM scorer.

\section{Experiments}
\label{sec:expriments}
In this section, we explore the performance of MultiLingPoT in simple and complex mathematical reasoning from both training and mixing perspectives.
In addition, we test MultiLingPoT on multiple models to examine its generalisability.

\newcolumntype{a}{>{\centering\arraybackslash}p{1cm}}
\newcolumntype{b}{>{\columncolor{Orange!10}\arraybackslash}l}
\newcolumntype{d}{>{\columncolor{Yellow!10}\centering\arraybackslash}p{1cm}}
\newcolumntype{e}{>{\columncolor{Green!7}\centering\arraybackslash}p{1cm}}
\newcolumntype{f}{>{\columncolor{Blue!7}\centering\arraybackslash}p{1cm}}
\begin{table*}[t]
  \centering
  \small
    \begin{tabular}{cbdeeeef}
    \toprule
    \rowcolor{white} \textbf{Method} & \textbf{Language} & \textbf{GSM8K} & \textbf{SVAMP} & \textbf{NumG} & \textbf{Mat} & \textbf{ASDiv} & \textbf{Average} \\
    \midrule
    & Python & 64.06 & 71.70 & 44.58 & \underline{54.07} & 72.60 & 61.40 \\
    & C++ & 64.97 & 71.50 & 43.30 & 33.08 & 71.83 & 56.93 \\
    & Java & 63.00 & 72.80 & 43.01 & 42.90 & 75.00 & 59.34 \\
    & Matlab & 62.62 & 69.70 & 42.30 & 34.55 & 71.83 & 56.20 \\
    \cmidrule{2-8}   
    \multirow{-5}{*}{SinglePoT} & Python-DA & 64.36 & 73.80 & 42.45 & 52.50 & \textbf{77.54} & 62.13 \\
    \midrule
    & Python & 65.57 & 73.10 & \underline{45.44} & 53.23 & 73.65 & \underline{62.19} \\
    & C++ & 64.97 & 73.60 & \underline{45.44} & 37.16 & 73.70 & 58.97 \\
    & Java & \underline{67.02} & \underline{75.10} & 44.87 & 45.19 & 73.80 & 61.19 \\
    & Matlab & 65.42 & 73.30 & 44.01 & 37.89 & 72.26 & 58.57 \\
    \cmidrule{2-8}    
    \rowcolor{Gray!10} \multirow{-5}{*}{\cellcolor{white} MultiLingPoT} & Self-Cons. & \textbf{69.37} & \textbf{75.60} & \textbf{46.72} & \textbf{54.69} & \underline{75.09} & \textbf{64.29} \\
    \bottomrule
    \end{tabular}%
  \caption{Results of MultiLingPoT training on simple datasets. ``-DA'' indicates data augmentation, and ``Self-Cons.'' refers to Self-Consistency. \textcolor{color1}{\rule{1.5ex}{1.5ex}} indicates in-domain testing, and \textcolor{color2}{\rule{1.5ex}{1.5ex}} indicates out-of-domain testing}
  \label{tab:sft-simple}%
\end{table*}%

\newcolumntype{a}{>{\centering\arraybackslash}p{1cm}}
\newcolumntype{b}{>{\columncolor{Orange!10}\arraybackslash}l}
\newcolumntype{d}{>{\columncolor{Yellow!10}\centering\arraybackslash}p{1cm}}
\newcolumntype{e}{>{\columncolor{Green!7}\centering\arraybackslash}p{1cm}}
\newcolumntype{f}{>{\columncolor{Blue!7}\centering\arraybackslash}p{1cm}}
\begin{table*}[t]
  \centering
  \small
    \begin{tabular}{cbdddddddf}
    \toprule
    \rowcolor{white} & \multicolumn{1}{c}{} &  & \textbf{Count} &  & \textbf{Int.} & \textbf{Num.} &  &  &  \\
    \rowcolor{white} \multirow{-2}{*}{\textbf{Method}} & \multicolumn{1}{c}{\multirow{-2}{*}{\textbf{Language}}} & \multirow{-2}{*}{\textbf{Algebra}} & \textbf{Prob.} & \multirow{-2}{*}{\textbf{Geom.}} & \textbf{Algebra} & \textbf{Theory} & \multirow{-2}{*}{\textbf{Prealg.}} & \multirow{-2}{*}{\textbf{Precalc.}} & \multirow{-2}{*}{\textbf{Average}} \\
    \midrule
    & Python & 35.94 & 24.89 & 17.69 & 15.55 & 48.17 & 45.51 & 21.54 & 29.89 \\
    & C++ & 37.85 & 27.42 & 21.10 & 16.73 & 39.73 & 46.59 & 19.37 & 29.82 \\
    & Java & 39.30 & 31.85 & 19.61 & 19.21 & 43.76 & 48.38 & 23.24 & 32.19 \\
    & Matlab & 29.48 & 34.59 & 18.12 & 11.50 & 40.69 & 43.72 & 19.61 & 28.24 \\
    \cmidrule{2-10}   
    \multirow{-5}{*}{SinglePoT} & Python-DA & 39.94 & 31.43 & 22.60 & \underline{21.56} & \underline{52.01} & 47.90 & \textbf{26.39} & \underline{34.54} \\
    \midrule
    & Python & 38.85 & 26.16 & 21.10 & 19.86 & 48.94 & 49.58 & 21.54 & 32.29 \\
    & C++ & \underline{41.12} & 30.80 & 21.53 & 18.95 & 42.80 & \underline{52.56} & 22.03 & 32.82 \\
    & Java & 40.94 & 33.33 & \underline{23.66} & 18.82 & 47.21 & 51.97 & 22.51 & 34.06 \\
    & Matlab & 31.11 & \underline{36.70} & 20.89 & 12.28 & 46.06 & 47.55 & 19.85 & 30.63 \\
    \cmidrule{2-10}   
    \rowcolor{Gray!10} \multirow{-5}{*}{\cellcolor{white} MultiLingPoT} & Self-Cons. & \textbf{45.04} & \textbf{37.76} & \textbf{23.88} & \textbf{23.13} & \textbf{53.16} & \textbf{57.34} & \underline{24.21} & \textbf{37.78} \\
    \bottomrule
    \end{tabular}%
  \caption{Results of MultiLingPoT training on complex datasets. ``-DA'' indicates data augmentation, and ``Self-Cons.'' refers to Self-Consistency.}
  \label{tab:sft-complex}%
\end{table*}%

\subsection{Experiments Setup}

\paragraph{Training Setup}
\label{sec:trainsetup}

We perform full fine-tuning of \texttt{CodeLlama-7B-hf}~\citep{roziere2023code} on the dataset constructed in Section~\ref{sec:dataconstruct} to obtain the MultiLingPoT model.
During training, we set the learning rate to 2e-5, the global batch size to 128, and the maximum sequence length to 1024 for three epochs.
To accelerate training, we use DeepSpeed ZeRO Stage 3~\citep{rajbhandari2020zero}.
All training operations are performed using Llama-Factory~\citep{zheng2024llamafactory}.

\paragraph{Evaluation Setup}

Since the difficulty of the problem affects the training and hybrid of MultiLingPoT, we test it on both simple and complex datasets.
For the simple dataset, we train on the GSM8K variant dataset from Section~\ref{sec:dataconstruct}, and additionally use SVAMP~\citep{patel2021nlp}, NumGLUE~\citep{mishra2022numglue}, Mathematics~\citep{davies2021advancing}, and ASDiv~\citep{miao2021diverse} for evaluation.
For the complex dataset, we train on the MATH variant dataset from Section~\ref{sec:dataconstruct} and test on seven categories from the MATH testset: Algebra, Counting \& Probability, Geometry, Intermediate Algebra, Number Theory, Prealgebra, and Precalculus.
All the evaluations use accuracy as the metric.

\paragraph{Baselines}

To assess the effectiveness of MultiLingPoT, we compare it with single-language PoT as follows:
1) SinglePoT in different languages: We train four SinglePoT models using training data from Python, C++, Java, and MATLAB, with each model performing reasoning in single programming language.
2) SinglePoT with data augmentation: To keep the total training data consistent, we augment the Python PoT data through multiple rounds of sampling, resulting in 26,414 samples for GSM8K and 14,634 samples for MATH.
Based on this augmented data, we train SinglePoT with data augmentation (i.e., SinglePoT-DA) as another baseline.

\subsection{MultiLingPoT Training Enhances Mathematical Reasoning}
\label{sec:exp1}

Table~\ref{tab:sft-simple} and Table~\ref{tab:sft-complex} show the performance of MultiLingPoT on simple and complex datasets after fine-tuning.

For the simple datasets, we have the following conclusions:
1) No single language is always optimal, which justifies the use of multiple languages rather than a single language to complete reasoning.
2) The training of MultiLingPoT enhances the reasoning performance across all languages.
Compared to SinglePoT, each language in MultiLingPoT improves about 2\%, which proves that different languages can also learn from each other.
3) Surprisingly, Python in MultiLingPoT consistently outperforms SinglePoT Python-DA.
This may be because, in simple problems, single-language augmentations have less diversity.
In contrast, solutions in different programming languages offer more diversity, which helps MultiLingPoT to perform better both within and outside the domain.

For the complex datasets, we find that:
1) The complex dataset more clearly reveals preferences for different programming languages.
For example, Python excels in Number Theory, Java performs well in Geometry, and MATLAB stands out in Counting \& Probability.
2) On complex datasets, MultiLingPoT training also improves the performance of each language.
However, since the complex problems have more diverse solutions, SinglePoT-DA still outperforms the individual languages of MultiLingPoT on some subsets.
But by simple self-consistency, MultiLingPoT outperform SinglePoT-DA by over 3\%.

\newcolumntype{a}{>{\centering\arraybackslash}p{1cm}}
\newcolumntype{b}{>{\columncolor{Orange!10}\arraybackslash}l}
\newcolumntype{d}{>{\columncolor{Yellow!10}\centering\arraybackslash}p{1cm}}
\newcolumntype{e}{>{\columncolor{Green!7}\centering\arraybackslash}p{1cm}}
\newcolumntype{f}{>{\columncolor{Blue!7}\centering\arraybackslash}p{1cm}}
\begin{table*}[t]
  \centering
  \small
    \begin{tabular}{cbdeeeef}
    \toprule
    \rowcolor{white} \multicolumn{2}{c}{\textbf{Method}} & \textbf{GSM8K} & \textbf{SVAMP} & \textbf{NumG} & \textbf{Mat} & \textbf{ASDiv} & \textbf{Average} \\
    \midrule
    & Python-DA & 64.36 & 73.80 & 42.45 & 52.50 & 77.54 & 62.13 \\
    \multirow{-2}{*}{SinglePoT} & Python-DA.SC & 67.24 & 74.00 & 43.58 & \underline{56.47} & \textbf{78.11} & 63.88 \\
    \midrule
    & Case-Based Choice & 65.35 & 73.40 & 45.01 & 50.93 & 74.23 & 61.78 \\
    & Bert Scorer & 66.18 & 74.10 & 44.15 & 42.58 & 73.75 & 60.15 \\
    & CodeBert Scorer & 65.88 & 73.30 & 44.30 & 37.99 & 72.21 & 58.73 \\
    & Llama3 Scorer & 64.82 & 73.70 & 45.44 & 36.95 & 72.93 & 58.76 \\
    \multirowcell{-5}{MultiLingPoT\\ Prior} & DPO & 62.85 & 70.70 & 42.87 & 39.56 & 71.83 & 57.56 \\ 
    \midrule
    & Self Consistency & 69.37 & 75.60 & 46.72 & 54.69 & 75.09 & 64.29 \\
    & Bert Scorer & 65.65 & 74.20 & 45.01 & 40.81 & 73.22 & 59.77 \\
    & CodeBert Scorer & 66.86 & 75.30 & 43.87 & 44.78 & 73.27 & 60.81 \\
    & Llama3 Scorer & \textbf{72.55} & \textbf{78.00} & \textbf{48.43} & 55.53 & \underline{77.59} & \textbf{66.42} \\
    \multirowcell{-5}{MultiLingPoT\\Posterior} & Voting \& Scorer & \underline{70.88} & \underline{77.90} & \underline{47.43} & \textbf{56.78} & 76.19 & \underline{65.83} \\
    \midrule 
    \rowcolor{Gray!10} \cellcolor{white} & Random & 64.82 & 72.50 & 44.30 & 43.31 & 72.88 & 59.56 \\
    \rowcolor{Gray!10} \cellcolor{white} & Upper Bound & 79.37 & 83.70 & 53.84 & 64.61 & 81.09 & 72.52 \\
    \bottomrule
    \end{tabular}%
  \caption{Results of MultiLingPoT with hybrid strategies on simple datasets. ``-DA'' indicates data augmentation, ``SC'' indicates Self Consistency. ``Random'' and ``Upper Bound'' are for MultiLingPoT and are provided as baselines.}
  \label{tab:hybrid-simple}%
\end{table*}%

\newcolumntype{a}{>{\centering\arraybackslash}p{0.9cm}}
\newcolumntype{b}{>{\columncolor{Orange!10}\arraybackslash}l}
\newcolumntype{d}{>{\columncolor{Yellow!10}\centering\arraybackslash}p{1cm}}
\newcolumntype{e}{>{\columncolor{Green!7}\centering\arraybackslash}p{1cm}}
\newcolumntype{f}{>{\columncolor{Blue!7}\centering\arraybackslash}p{1cm}}
\begin{table*}[t]
  \centering
  \small
    \begin{tabular}{cbdddddddf}
    \toprule
    \rowcolor{white} \multicolumn{2}{c}{} &  & \textbf{Count} &  & \textbf{Int.} & \textbf{Num.} &  &  &  \\
    \rowcolor{white} \multicolumn{2}{c}{\multirow{-2}{*}{\textbf{Method}}} & \multirow{-2}{*}{\textbf{Algebra}} & \textbf{Prob.} & \multirow{-2}{*}{\textbf{Geom.}} & \textbf{Algebra} & \textbf{Theory} & \multirow{-2}{*}{\textbf{Prealg.}} & \multirow{-2}{*}{\textbf{Precalc.}} & \multirow{-2}{*}{\textbf{Average}} \\
    \midrule
    & Python-DA & 39.94 & 31.43 & 22.60 & 21.56 & 52.01 & 47.90 & \underline{26.39} & 34.54 \\
    \multirow{-2}{*}{SinglePoT} & Python-DA.SC & 44.94 & 33.54 & 23.02 & 25.22 & \underline{56.04} & 52.80 & \textbf{26.63} & 37.45 \\
    \midrule
    & Case-Based & 43.03 & 33.12 & 22.60 & 21.56 & 48.75 & 52.44 & 24.45 & 35.13 \\
    & Bert Scorer & 38.76 & 35.65 & 22.60 & 20.39 & 51.24 & 51.73 & 21.54 & 34.55 \\
    & CodeBert Scorer & 40.40 & 34.59 & 21.74 & 21.30 & 49.71 & 51.85 & 23.97 & 34.79 \\
    & Llama3 Scorer & 36.94 & 31.22 & 21.10 & 18.56 & 45.48 & 53.64 & 21.54 & 32.64 \\
    \multirowcell{-5}{MultiLingPoT\\Prior} & DPO & 33.57 & 36.28 & 17.91 & 18.16 & 45.10 & 42.17 & 20.33 & 30.50 \\
    \midrule
    & Self Consistency & \underline{45.04} & 37.76 & 23.88 & 23.13 & 53.16 & 57.34 & 24.21 & 37.78 \\
    & Bert Scorer & 43.22 & 35.86 & 22.17 & 23.39 & 51.24 & 52.92 & 23.00 & 35.97 \\
    & CodeBert Scorer & 39.67 & 35.65 & 23.45 & 23.13 & 49.52 & 53.64 & 23.24 & 35.47 \\
    & Llama3 Scorer & 41.40 & \underline{41.98} & \underline{24.09} & \underline{25.88} & 55.47 & \underline{59.49} & 24.93 & \underline{39.03} \\
    \multirowcell{-5}{MultiLingPoT\\Posterior} & Voting \& Scorer & \textbf{47.95} & \textbf{42.40} & \textbf{24.52} & \textbf{26.01} & \textbf{57.00} & \textbf{59.73} & 26.15 & \textbf{40.53} \\
    \midrule
    \rowcolor{Gray!10} \cellcolor{white} & Random & 34.03 & 30.16 & 22.60 & 16.99 & 45.87 & 51.37 & 21.30 & 31.76 \\
    \rowcolor{Gray!10} \cellcolor{white} & Upper Bound & 58.32 & 53.37 & 34.11 & 34.37 & 68.13 & 68.45 & 33.89 & 50.09 \\
    \bottomrule
    \end{tabular}%
  \caption{Results of MultiLingPoT with hybrid strategies on complex datasets. ``-DA'' indicates data augmentation, ``SC'' indicates Self Consistency. ``Random'' and ``Upper Bound'' are for MultiLingPoT and are provided as baselines.}
  \label{tab:hybrid-complex}%
\end{table*}%

\newcolumntype{a}{>{\centering\arraybackslash}p{1cm}}
\newcolumntype{b}{>{\columncolor{Orange!10}\arraybackslash}l}
\newcolumntype{d}{>{\columncolor{Yellow!10}\centering\arraybackslash}p{1cm}}
\newcolumntype{e}{>{\columncolor{Green!7}\centering\arraybackslash}p{1cm}}
\newcolumntype{f}{>{\columncolor{Blue!7}\centering\arraybackslash}p{1cm}}
\begin{table*}[t]
  \centering
  \small
    \begin{tabular}{cbdddddddf}
    \toprule
    \rowcolor{white} & \multicolumn{1}{c}{} &  & \textbf{Count} &  & \textbf{Int.} & \textbf{Num.} &  &  &  \\
    \rowcolor{white} \multirow{-2}{*}{\textbf{Method}} & \multicolumn{1}{c}{\multirow{-2}{*}{\textbf{Language}}} & \multirow{-2}{*}{\textbf{Algebra}} & \textbf{Prob.} & \multirow{-2}{*}{\textbf{Geom.}} & \textbf{Algebra} & \textbf{Theory} & \multirow{-2}{*}{\textbf{Prealg.}} & \multirow{-2}{*}{\textbf{Precalc.}} & \multirow{-2}{*}{\textbf{Average}} \\
    \midrule
    \multicolumn{10}{c}{DeepseekCoder} \\
    \midrule
    SinglePoT & Python-DA & 49.40 & 37.97 & 26.01 & \underline{27.18} & \underline{61.03} & 62.96 & \underline{27.36} & \underline{41.70} \\
    \midrule
    & Python & 47.95 & 31.64 & 25.37 & 23.92 & 60.26 & 60.81 & 26.87 & 39.54 \\
    & C++ & \underline{53.95} & 41.56 & 27.07 & 22.74 & 52.20 & \underline{64.39} & 26.39 & 41.18 \\
    & Java & 48.77 & \underline{43.67} & \underline{29.42} & 23.79 & 54.70 & 62.72 & 26.87 & 41.42 \\
    & Matlab & 35.21 & 43.45 & 26.86 & 14.50 & 51.82 & 56.15 & 24.45 & 36.06 \\
    \cmidrule{2-10}
    \rowcolor{Gray!10} \multirow{-5}{*}{\cellcolor{white} MultiLingPoT} & Self-Cons. & \textbf{58.23} & \textbf{49.57} & \textbf{30.49} & \textbf{28.36} & \textbf{62.95} & \textbf{68.21} & \textbf{30.02} & \textbf{46.83} \\
    \midrule
    \multicolumn{10}{c}{CodeLlama-Python} \\
    \midrule
    SinglePoT & Python-DA & \underline{40.85} & 32.91 & \underline{23.45} & \underline{21.04} & 52.59 & \underline{52.44} & 23.97 & \underline{35.32} \\
    \midrule
    & Python & 40.30 & 28.48 & 18.97 & 19.86 & \underline{53.93} & 50.77 & 23.72 & 33.71 \\
    & C++ & 40.76 & 32.70 & 23.24 & 18.43 & 47.79 & 50.65 & 21.30 & 33.55 \\
    & Java & 39.67 & 36.91 & 20.46 & 18.69 & 48.94 & 52.09 & \underline{24.45} & 34.45 \\
    & Matlab & 30.48 & \underline{38.18} & 21.10 & 13.20 & 45.68 & 47.31 & 22.76 & 31.24 \\
    \cmidrule{2-10}
    \rowcolor{Gray!10} \multirow{-5}{*}{\cellcolor{white} MultiLingPoT} & Self-Cons. & \textbf{46.95} & \textbf{40.08} & \textbf{24.09} & \textbf{23.26} & \textbf{56.81} & \textbf{57.10} & \textbf{26.87} & \textbf{39.30} \\
    \midrule
    \multicolumn{10}{c}{Llama3} \\
    \midrule
    SinglePoT & Python-DA & 42.12 & 30.80 & 23.66 & \textbf{24.44} & \textbf{55.85} & 52.21 & 21.79 & \underline{35.83} \\
    \midrule
    & Python & 40.49 & 28.05 & 21.10 & 17.90 & 48.94 & \underline{54.24} & 20.82 & 33.07 \\
    & C++ & \underline{44.13} & 34.59 & 23.45 & 19.60 & 45.87 & 53.28 & \underline{22.27} & 34.74 \\
    & Java & 44.04 & 38.81 & \underline{25.79} & 19.21 & 46.25 & 53.88 & 21.79 & 35.68 \\
    & Matlab & 30.11 & \underline{40.08} & 20.46 & 13.33 & 43.76 & 46.35 & 21.06 & 30.73 \\
    \cmidrule{2-10}
    \rowcolor{Gray!10} \multirow{-5}{*}{\cellcolor{white} MultiLingPoT} & Self-Cons. & \textbf{50.22} & \textbf{41.98} & \textbf{26.22} & \underline{23.92} & \underline{54.70} & \textbf{59.73} & \textbf{25.66} & \textbf{40.34} \\
    \bottomrule
    \end{tabular}%
  \caption{Results of different models using MultiLingPoT on complex datasets, 
  including DeepseekCoder for the code model, CodeLlama-Python for the code model in a single language, and Llama3 for the non-code model.
  }
  \label{tab:more-models}%
\end{table*}%

\subsection{Hybrid Boosts MultiLingPoT}

Table~\ref{tab:hybrid-simple} and Table~\ref{tab:hybrid-complex} show the performance of MultiLingPoT on the simple and complex datasets with different hybrid strategies.

On the simple datasets, MultiLingPoT outperforms SinglePoT-DA by 4\% through mixing, and also surpasses SinglePoT-DA with self-consistency.
Comparing the two mixing strategies, the performance of the posterior hybrid far exceeds the priori hybrid.
This may be because the posterior approach relies on both the query and the generated program's correctness.
The prior approach performs similarly to random selection, suggesting that judging the language only based on the query is still challenging for the simple dataset.
Among the hybrid implementations, the Llama3 Scorer in the posterior approach performs best, as LLMs excel at understanding and judging compared to smaller models like Bert.
Finally, DPO performs the worst of all the hybrid implementations, tending to choose the same language for most of the questions and lacking in preference selection.
This may be due to the use of different languages under the same prompt template leading to confusion in the model and ultimately the selection of a single language.
In addition, due to the small training set of GSM8K, the lack of training data for DPO also affected its performance.

In complex datasets, MultiLingPoT also outperforms SinglePoT-DA and brings greater improvement.
Similar to the simple datasets, the posterior hybrid strategy also better than the prior hybrid in complex datasets.
However, compared to random selection, the prior hybrid still shows some improvement.
This may be because language preferences are more pronounced in complex questions, as noted by Section~\ref{sec:exp1}.
Finally, In terms of specific hybrid implementations, Voting \& Scorer performs best.
This is because self-consistency often leads to ties in the complex dataset, allowing the Scorer to play a more significant role.
Additionally, the errors in complex questions are more diverse, so voting is less likely to cause incorrect results.

\subsection{Different Models on MultiLingPoT}

To further explore the applicability of MultiLingPoT, we repeat the method on other models.
Since the hybrid strategy is independent of the model, we focus on the impact of different models on MultiLingPoT training.
We choose three types of models: 
\texttt{DeepseekCoder-7B-v1.5} for other code models~\citep{zhu2024deepseek}, \texttt{CodeLlama-7B-Python-hf} for single-language code models~\citep{roziere2023code} and \texttt{Llama3-8B-Instruct} for non-code models~\citep{llama3modelcard}.
Table~\ref{tab:more-models} shows the results of different models using MultiLingPoT on complex dataset.

Among these three types of models, MultiLingPoT with self-consistency consistently preserves the best results, and the performance of each language is generally balanced, demonstrating the broad applicability of the MultiLingPoT in models with code capabilities.
Certainly, there are still differences between the different models.
For DeepseekCoder, both its SinglePoT and MultiLingPoT perform better, indicating that the performance of MultiLingPoT improves with the model's inherent capabilities.
For Llama3, the performance is only matched by CodeLlama based on Llama2, indicating that the code capability of the model still affects the effectiveness of MultiLingPoT.
Finally, CodeLlama-Python performs slightly better in Python, indicating that language-specific models can improve performance in their focus language without significantly affecting others.

\section{Conclusion}
\label{sec:conclusion}
In this paper, we propose MultiLingPoT, a scalable and automated method to enhance the reasoning capabilities of LLM via multilingual program-of-thought.
% multilingual program reasoning method based on SFT.
Using GSM8K and MATH, we construct multi-language simple and complex datasets with 26,359 and 14,775 samples.
By fine-tuning on the multi-language dataset, we obtain the MultiLingPoT model, which can answer questions in multiple languages.
Further, we explore hybrid methods categorized into prior and posterior, allowing the model to answer the corresponding question using the most appropriate language.
Our experimental results show that:
1) In the training phase, the boost of simple questions is more pronounced, as the diversity gain from simple question augmentation is less than that from the variability between different languages.
2) In the hybrid phase, complex questions exhibit greater potential, as they reveal more pronounced language preference differences.
3) Overall, after training and mixing, MultiLingPoT outperforms single-language PoT in both simple and complex questions, delivering about 6\% improvement, and is widely applicable to models with code capabilities.

\section*{Limitations}
\label{sec:limitation}
Our study is comprehensive but still has some limitations that we plan to address in future research.
For the multilingual PoT data we construct, we use ChatGPT, as the work is conducted at that time.
Although more advanced models, such as GPT-4~\citep{openai2023gpt4}, are now available and can generate higher-quality data, we do not rebuild the dataset due to time and budget constraints.
In addition, there is room for improvement for hybrid strategies.
Although we conduct many explorations, we still not find an a prior hybrid method that is obviously effective.
While posterior hybrid is effective, it is more computationally intensive for each problem.
Therefore, it is still a challenge to investigate how to effectively mix different programming languages while minimising the amount of computation required.
Finally, we currently focuse only on the application of multiple programming languages to mathematical problems.
Since PoT has been widely applied in various domains, we will also explore the use of multi-language PoT in other fields in the future.

\section*{Ethics Considerations}
\label{sec:Ethics}
We hereby acknowledge that all authors of this work are aware of the provided ACL Code of Ethics and honor the code of conduct.
In our work, we use publicly available data during the dataset construction process and perform secondary construction.
We strictly follow the ChatGPT usage guidelines and performed subsequent validation of the generated content to minimise the risk of harmful content generation.
During model training, we adopt a publicly standardised training process and used harmless datasets to safeguard the model.

\bibliography{custom}

\clearpage
\appendix
\begin{appendix}
\label{sec:appendix}
\section{Prompt Template}

\subsection{Prompt Template for Data Construction}
\label{appendix:data-construction}
The prompt template for ChatGPT to generate the multilingual PoT data is shown in List~\ref{lst:data-construction}.
\lstset{
    backgroundcolor=\color[RGB]{245,245,244},
    breaklines=true,
    breakindent=0pt,
    basicstyle=\ttfamily\small,
    escapeinside={(*@}{@*)},
    emph={Task,Prompt,Example,Python,Java,C++,Matlab,MATH,GSM8K, Questions,Solutions},
    emphstyle={\bfseries\color{NavyBlue}}
}\begin{lstlisting}[caption={Instruction template for ChatGPT to generate multilingual PoT data.},label=lst:data-construction]
Task Prompt:
Please use {program_type} functions to solve math problems. The function name is "solution()" and return the result. The following are some cases:

Example Questions of GSM8K:
Question1: Natalia sold clips to 48 of her friends in April, and then she sold half as many clips in May. How many clips did Natalia sell altogether in April and May?

Question2: There are 381 pages in Elliot book.  He has already read 149 pages.  If he reads 20 pages a day for a week, how many pages are still left to be read?

Question3: Weng earns $12 an hour for babysitting. Yesterday, she just did 50 minutes of babysitting. How much did she earn?

Question4: Alexis is applying for a new job and bought a new set of business clothes to wear to the interview. She went to a department store with a budget of $200 and spent $30 on a button-up shirt, $46 on suit pants, $38 on a suit coat, $11 on socks, and $18 on a belt. She also purchased a pair of shoes, but lost the receipt for them. She has $16 left from her budget. How much did Alexis pay for the shoes?

Example Solutions of Python in GSM8K:
def solution():
    clips_april = 48
    clips_may = clips_april / 2
    clips_total = clips_april + clips_may
    result = clips_total
    return result

def solution():
    pages_initial = 381
    pages_read = 149
    pages_per_day = 20
    num_days = 7  # 7 days in a week
    pages_read_in_week = pages_per_day * num_days
    pages_left = pages_initial - pages_read - pages_read_in_week
    result = pages_left
    return result

def solution():
    hourly_rate = 12
    minutes_worked = 50
    hours_worked = minutes_worked / 60
    earnings = hourly_rate * hours_worked
    result = earnings
    return result

def solution():
    budget = 200
    shirt = 30
    pants = 46
    coat = 38
    socks = 11
    belt = 18
    money_left = 16
    shoes = budget - (shirt + pants + coat + socks + belt + money_left)
    result = shoes
    return result
    
Example Solutions of (*@\textbf{\color{NavyBlue}C++}@*) in GSM8K:
float solution() {
    float clips_april = 48;
    float clips_may = clips_april / 2;
    float clips_total = clips_april + clips_may;
    float result = clips_total;
    return result;
}

float solution() {
    float pages_initial = 381;
    float pages_read = 149;
    float pages_per_day = 20;
    float num_days = 7;  // 7 days in a week
    float pages_read_in_week = pages_per_day * num_days;
    float pages_left = pages_initial - pages_read - pages_read_in_week;
    float result = pages_left;
    return result;
}

float solution() {
    float hourly_rate = 12;
    float minutes_worked = 50;
    float hours_worked = minutes_worked / 60;
    float earnings = hourly_rate * hours_worked;
    float result = earnings;
    return result;
}

float solution() {
    float budget = 200;
    float shirt = 30;
    float pants = 46;
    float coat = 38;
    float socks = 11;
    float belt = 18;
    float money_left = 16;
    float shoes = budget - (shirt + pants + coat + socks + belt + money_left);
    float result = shoes;
    return result;
}

Example Solutions of Java in GSM8K:
public static double solution() {
    double clips_april = 48;
    double clips_may = clips_april / 2;
    double clips_total = clips_april + clips_may;
    double result = clips_total;
    return result;
}

public static double solution() {
    double pages_initial = 381;
    double pages_read = 149;
    double pages_per_day = 20;
    double num_days = 7;  // 7 days in a week
    double pages_read_in_week = pages_per_day * num_days;
    double pages_left = pages_initial - pages_read - pages_read_in_week;
    double result = pages_left;
    return result;
}

public static double solution() {
    double hourly_rate = 12;
    double minutes_worked = 50;
    double hours_worked = minutes_worked / 60;
    double earnings = hourly_rate * hours_worked;
    double result = earnings;
    return result;
}

public static double solution() {
    double budget = 200;
    double shirt = 30;
    double pants = 46;
    double coat = 38;
    double socks = 11;
    double belt = 18;
    double money_left = 16;
    double shoes = budget - (shirt + pants + coat + socks + belt + money_left);
    double result = shoes;
    return result;
}

Example Solutions of Matlab in GSM8K:
function result = solution()
    clipsApril = 48;
    clipsMay = clipsApril / 2;
    totalClips = clipsApril + clipsMay;
    result = totalClips;
end

function result = solution()
    totalPages = 381;
    pagesRead = 149;
    pagesPerDay = 20;
    daysInAWeek = 7;
    pagesInAWeek = pagesPerDay * daysInAWeek;
    remainingPages = totalPages - pagesRead - pagesInAWeek;
    result = remainingPages;
end

function result = solution()
    hourlyRate = 12;
    babysittingMinutes = 50;
    babysittingHours = babysittingMinutes / 60;
    earnings = hourlyRate * babysittingHours;
    result = earnings;
end

function result = solution()
    budget = 200;
    shirtCost = 30;
    pantsCost = 46;
    coatCost = 38;
    socksCost = 11;
    beltCost = 18;
    amountSpent = shirtCost + pantsCost + coatCost + socksCost + beltCost;
    remainingBudget = budget - amountSpent;
    shoesCost = budget - amountSpent;
    result = shoesCost;
end

Example Questions of MATH:
Question1: The function $f(x)$ satisfies [f(x + y) = f(x) f(y)]for all real numbers $x$ and $y.$ If $f(2) = 3,$ find $f(6).$

Question2: Compute the sum of all the roots of $(2x+3)(x-4)+(2x+3)(x-6)=0$.

Question3: A triangle in a Cartesian coordinate plane has vertices (5, -2), (10, 5) and (5, 5). How many square units are in the area of the triangle? Express your answer as a decimal to the nearest tenth

Question4: How many nonnegative solutions are there to the equation $x^2 = -4x$?

Example Solutions of Python in MATH:
def solution():
    def f(x):
        if x == 2:
            return 3
        else:
            return f(2) * f(x - 2)
    result = f(6)
    return result

def solution():
    x = sp.symbols('x')
    equation = (2*x + 3)*(x - 4) + (2*x + 3)*(x - 6)
    roots = sp.solve(equation, x)
    result = sum(roots)
    return result

def solution():
    vertices = [(5, -2), (10, 5), (5, 5)]
    area = 0.5 * abs((vertices[0][0]*(vertices[1][1]-vertices[2][1]) + vertices[1][0]*(vertices[2][1]-vertices[0][1]) + vertices[2][0]*(vertices[0][1]-vertices[1][1])))
    return round(area, 1)

import sympy as sp
def solution():
    x = sp.symbols('x')
    equation = x**2 + 4*x
    solutions = sp.solve(equation, x)
    non_negative_solutions = [sol for sol in solutions if sol >= 0]
    result = len(non_negative_solutions)
    return result

Example Solutions of (*@\textbf{\color{NavyBlue}C++}@*) in MATH:
double f(double x) {
    if (x == 2) {
        return 3;
    }
    return f(2) * f(x - 2);
}
double solution() {
    return f(6);
}

double solution() {
    double root1 = -1.5;
    double root2 = 5;
    double sum_of_roots = root1 + root2;
    return sum_of_roots;
}

double solution() {
    double x1 = 5.0, y1 = -2.0;
    double x2 = 10.0, y2 = 5.0;
    double x3 = 5.0, y3 = 5.0;
    double area = 0.5 * std::abs(x1 * (y2 - y3) + x2 * (y3 - y1) + x3 * (y1 - y2));
    return area;
}

int solution() {
    int a = 1;
    int b = 4;
    int discriminant = b * b - 4 * a * 0;
    int root_count = 0;
    if (discriminant > 0) {
        double root1 = (-b + sqrt(discriminant)) / (2 * a);
        double root2 = (-b - sqrt(discriminant)) / (2 * a);
        if (root1 >= 0) root_count++;
        if (root2 >= 0) root_count++;
    } else if (discriminant == 0) {
        double root = -b / (2 * a);
        if (root >= 0) root_count++;
    }
    return root_count;
}

Example Solutions of Java in MATH:
public static double solution() {
    double a = Math.sqrt(3);
    double result = Math.pow(a, 6);
    return result;
}

public static double solution() {
    double a = 4;
    double b = -14;
    double c = -30;
    double discriminant = b * b - 4 * a * c;
    if (discriminant < 0) {
        return Double.NaN;
    } else {
        double root1 = (-b + Math.sqrt(discriminant)) / (2 * a);
        double root2 = (-b - Math.sqrt(discriminant)) / (2 * a);

        double sumOfRoots = root1 + root2;

        return sumOfRoots;
    }
}

public static double solution() {
    double x1 = 5;
    double y1 = -2;
    double x2 = 10;
    double y2 = 5;
    double x3 = 5;
    double y3 = 5;
    double area = 0.5 * Math.abs(x1 * y2 + x2 * y3 + x3 * y1 - x1 * y3 - x2 * y1 - x3 * y2);
    return area;
}

public static int solution() {
    int a = 1;
    int b = 4;
    int discriminant = b * b - 4 * a * 0;
    int root_count = 0;
    if (discriminant > 0) {
        double root1 = (-b + Math.sqrt(discriminant)) / (2 * a);
        double root2 = (-b - Math.sqrt(discriminant)) / (2 * a);
        if (root1 >= 0) root_count++;
        if (root2 >= 0) root_count++;
    } else if (discriminant == 0) {
        double root = -b / (2.0 * a);
        if (root >= 0) root_count++;
    }
    return root_count;
}

Example Solutions of Matlab in MATH:
function result = solution()
    function value = f(x)
        if x == 2
            value = 3;
        else
            value = f(x - 2) * f(2);
        end
    end
    result = f(6);
end

function result = solution()
    syms x;
    equation = (2*x + 3)*(x - 4) + (2*x + 3)*(x - 6) == 0;
    roots_x = solve(equation, x);
    sum_of_roots = sum(roots_x);
    result = sum_of_roots;
end

function result = solution()
    x1 = 5; y1 = -2;
    x2 = 10; y2 = 5;
    x3 = 5; y3 = 5;
    area = 0.5 * abs(x1*(y2 - y3) + x2*(y3 - y1) + x3*(y1 - y2));
    result = area;
end

function result = solution()
    syms x
    equation = x^2 + 4*x;
    solutions = solve(equation);
    num_solutions = sum(double(solutions >= 0));
    result = num_solutions;
end

\end{lstlisting}

\subsection{Prompt Template for Model Inference}
\label{appendix:model-inference}
The prompt template for MultiLingPoT to inference in different languages is shown in List~\ref{lst:model-inference}.
\lstset{
    backgroundcolor=\color[RGB]{245,245,244},
    breaklines=true,
    breakindent=0pt,
    basicstyle=\ttfamily\small,
    escapeinside={(*@}{@*)},
    emph={Instruction, Input, Output, Python, C++, Java, Matlab},
    emphstyle={\bfseries\color{NavyBlue}}
}\begin{lstlisting}[caption={Instruction template for  MultiLingPoT to inference in different languages.},label=lst:model-inference]
/* Instruction */
You are a math problem solving robot, for the input math problem, write the solution function to solve the problem using the specified programming language.

/* Input */
Write a {program_type} program to solve the problem: {question}

/* Output in Python */
def solution():
    x, y = sp.symbols('x y')
    equation1 = x**2 - y
    equation2 = x + y - 1
    intersections = sp.solve((equation1, equation2), (x, y))
    distance = sp.sqrt((intersections[0][0] - intersections[1][0])**2 + (intersections[0][1] - intersections[1][1])**2)
    return distance

/* Output in (*@\textbf{\color{NavyBlue}C++}@*) */
double solution() {
    // The equation of the first intersection is (0, 0)
    // The equation of the second intersection is (1 - sqrt(1 - 4*0), 1 - sqrt(1 - 4*0))
    double x1 = 0;
    double y1 = 0;
    double x2 = 1 - sqrt(1 - 4 * 0);
    double y2 = 1 - sqrt(1 - 4 * 0);
    double distance = sqrt(pow(x2 - x1, 2) + pow(y2 - y1, 2));
    return distance;
}

/* Output in Java */
public static double solution() {
    double x1 = 0;
    double y1 = 0;
    double x2 = 1 - Math.sqrt(1 - 4 * 0 * 0);
    double y2 = 1 - 0;
    double distance = Math.sqrt(Math.pow(x2 - x1, 2) + Math.pow(y2 - y1, 2));
    return distance;
}

/* Output in Matlab */
function result = solution()
    syms x;
    y = x^2;
    equation1 = x + y == 1;
    equation2 = y == 0;
    solutions = solve([equation1, equation2], x);
    intersection_points = [double(solutions(1)), 0];
    distance = norm(intersection_points(1) - intersection_points(2));
    result = distance;
end

\end{lstlisting}
\end{appendix}

\end{document}